\documentclass[nohyperref]{article}

\usepackage[accepted]{icml2023}

\usepackage[
backend=biber,
style=alphabetic,
sorting=ynt
]{biblatex}

\usepackage{amsmath}
\usepackage{amssymb}
\usepackage{mathtools}
\usepackage{amsthm}
\usepackage{graphicx}
\usepackage{subcaption}
\usepackage[capitalize,noabbrev]{cleveref}
\theoremstyle{plain}

\theoremstyle{definition}

\theoremstyle{remark}

\icmltitlerunning{Published as a conference paper at ICML 2023: Structured Probabilistic Inference and Generative Modeling}

\begin{document}

\twocolumn[
\icmltitle{Geometric Constraints in Probabilistic Manifolds: A Bridge from Molecular Dynamics to Structured Diffusion Processes}



\icmlsetsymbol{equal}{*}

\begin{icmlauthorlist}
\icmlauthor{Justin Diamond}{yyy}

\icmlauthor{Markus Lill}{yyy}
\end{icmlauthorlist}

\icmlaffiliation{yyy}{Department of Pharmaceutical Sciences, University of Basel, Basel, Switzerland}

\icmlcorrespondingauthor{Justin Diamond}{justin.diamond@unibas.ch}

\icmlkeywords{Machine Learning, ICML}

\vskip 0.3in
]



\printAffiliationsAndNotice{}  % leave blank if no need to mention equal contribution

\begin{abstract}

Understanding the macroscopic characteristics of biological complexes demands precision and specificity in statistical ensemble modeling. One of the primary challenges in this domain lies in sampling from particular subsets of the state-space, driven either by existing structural knowledge or specific areas of interest within the state-space.
We propose a method that enables sampling from distributions that rigorously adhere to arbitrary sets of geometric constraints in Euclidean spaces. This is achieved by integrating a constraint projection operator within the well-regarded architecture of Denoising Diffusion Probabilistic Models, a framework founded in generative modeling and probabilistic inference.
The significance of this work becomes apparent, for instance, in the context of deep learning-based drug design, where it is imperative to maintain specific molecular profile interactions to realize the desired therapeutic outcomes and guarantee safety. 
\end{abstract}

\section{Introduction}
Infinitesimal Dynamics in classical mechanics is commonly formalized by Lagrangians.
By solving for functionals that extremize the Lagrangian one obtains equations of motion. In molecular systems, e.g. Molecular Dynamics, the EOM  are:
 $M\frac{d^2x}{dt^2}=-\nabla{U} - \sum_a{\lambda_a\nabla{\sigma_a}}$,
where $M$ is the diagonal mass matrix, $x$ the cartesian coordinates, $t$ is time, 
and $U$ is the potential energy. The $\sigma_a$ are a set of holonomic constraints and $\lambda_a$ are the Lagrange multiplier coefficients. To generalize from holonomic to nonholonomic constraints, one can use slack variables to transform the latter into the first.

  Starting with $z_x, z_h = f(x,h) = [x(0), h(0)] +  \int_{0}^{1} \phi(x(t), h(t))dt$  with $z$ being a latent vector sampled from Gaussians and the indexes x and h indicate the latent variables associated
to the coordinates of each particle and the vector embedding of each particle, $\phi$ is the parameterized transformation defined by a equivariant graph neural network. This defines a Neural ODE \cite{chen2018neural} which generalizes to Denoising Diffusion Probabilistic Models \cite{ho2020denoising}. 
This form of transformation has the same infinitesimal nature as our previous EOM which makes it acceptable to apply sets of constraints via Langrange's Multipliers, analogous to solving our EOM and thus one can insure the continual satisfaction of a set of constraints
using the Shake algorithm from Molecular Dynamics. 

The study of constrained dynamics in Molecular Dynamics and Machine Learning, has traditionally focused on mostly linear constraints: e.g. removing high-frequency oscillations by constraining bond distances in the first and in-painting in the latter by thresholding certain pixel values to predetermined values. From a high level these can be seen as linear constraint problems as the constrained subset affects the unconstrained subset to minimal degrees. In addition, our task is more challenging as different constraints induce different geometric topological structures, such that some sets of distance constraints can determine uniquely the solution, and small modifications in the constraints may lead to vast changes in the solution set.  

The problem we hope to model are non-linear constraints where constrained subsets of atoms determine the unconstrained subset to a high degree. We argue these types of non-linear constraints are important in the field of generative drug development where generated molecules  must satisfy certain structural or analytic properties a priori. Take for instance, the optimization of lead molecules which is crucial at the final stages of drug development pipeline where off target interactions are attempted to be minimized. Since these off-target reactions can often be described by structural or analytic properties, then we can generate precisely molecules that satisfy a constraint profile of the target of interest, while specifying the subspace of generated molecules to not lie within the subspace of off-target interaction profiles. 

In the following, we will give a summary of the Shake algorithm and segments of the equivariant normalaizing flow necessary to elaborate on how to combine them. Next, it will be elaborated that the spaces of latent embeddings and output samples are generally of very different nature, and constraints defined in one space will not necessarily be useful in the other. We suggest a continuous transformation of
the constraints such that they are always satisfied in the latent space, and become more restrictive throughout the integration. Lastly, we show simple examples where complex constraints are satisfied within small molecules. We leave to future work the study of this methodology to larger systems, and more application based studies. Our approach builds a fruitful junction where probabilistic inference, structured data representation, and generative modeling meet, while emphasizing the necessity to encode domain knowledge effectively in these settings, offering a way to formally verify the distributions from which samples are drawn.
\section{Previous Research}

Generative models of graphs have been a subject of interest in recent years. A number of different approaches have been proposed in the literature. \cite{hoffmann2019generating} generates valid Euclidean distance matrices ensuring the resulting molecular structures are physically realistic which are then reconstructed in 3D space. In \cite{noe2019boltzmann}, Boltzmann Generators sample equilibrium states of many-body systems with deep learning, useful for generating molecular configurations that obey thermodynamics distributions.

\cite{satorras2021equivariant} proposed Equivariant Graph Neural Networks, which can be applied to model molecules and proteins while ensuring that their predictions are consistent under different orientations and permutations of the molecule.\cite{hoogeboom2023equivariant} further extended the concept to the diffusion process for 3D molecule generation. \cite{corso2023diffdock} applied similar methodologies to diffusion models on protein ligand complexes, and \cite{jing2023eigenfold} devise a method of protein generation models that diffuse over harmonic potentials.

The Shake algorithm, described in a parallelized fashion by \cite{elber2011shake}, enforces linear constraints on molecular dynamics simulations of chemicals and biomolecules. This algorithm is conventionally used in simulations to get rid of high frequency motions, i.e. those seen in bonds between atoms.

\section{Constrained Generative Processes}
\subsection{Geometric Constraints in Shake}
First, we define the constraint functions for the pairwise distance (not necessarily between bonded atoms), bond angle, and dihedral angle.

\begin{equation}
\sigma_{d_{ij}} = \left(d_{ij} - d_{ij,0}\right)^2 = 0 
\end{equation}
\begin{equation}
\sigma_{\theta_{ijk}} = \left(\theta_{ijk} - \theta_{ijk,0}\right)^2 = 0 
\end{equation}
\begin{equation}
\sigma_{\psi_{ijkl}} = \left(\psi_{ijkl} - \psi_{ijkl,0}\right)^2 = 0
\end{equation}

These constraint functions compare the current pairwise distance, bond angle, and dihedral angle with their target values, and the goal is to minimize the difference.  We can additionally create nonholonomic constraints via slack variables. For example, we can add a slack variable $y \geq 0$ and define $d_j$ as the boundary of a nonholonomic constraint. Then, we can express the constraint as:

$
\sigma_a := ||x_{aj} - x_{ak}||^2_2 - d_j \leq 0 \rightarrow ||x_{aj} - x_{ak}||^2_2 - d_j + y= 0$. 

Next, modify the constraint matrix in the Shake algorithm to include pairwise distance, bond angle, and dihedral angle constraints seen in equation 4, where $ij$, $ijk$, and $ijkl$ sum over the pairwise, bond angles, and torsion constraints indicating the number of atoms in each type of constraint type. 
\begin{figure*}
\begin{equation}
A^{(n-1)}_{\alpha\beta} = \frac{\partial^2 U}{\partial x_\alpha \partial x_\beta} +  \sum_{ij}\lambda^{(n-1)}{d_{ij}}\frac{\partial^2 \sigma_{d_{ij}}}{\partial x_\alpha \partial x_\beta} + \sum_{ijk}\lambda^{(n-1)}{\theta_{ijk}}\frac{\partial^2 \sigma_{\theta_{ijk}}}{\partial x_\alpha \partial x_\beta} + \sum_{ijkl}\lambda^{(n-1)}{\psi_{ijkl}}\frac{\partial^2 \sigma_{\psi_{ijkl}}}{\partial x_\alpha \partial x_\beta}
\end{equation}
\end{figure*}
The constraint matrix now accounts for the pairwise distance, bond angle, and dihedral angle constraints by including their second-order derivatives with respect to the Cartesian coordinates by including their contributions to the Lagrange multipliers. After solving for the Lagrange multipliers, update the coordinates using the adjusted coordinate set equation like before. 
It is also possible to try to optimize the coordinates via other optimization algorithms like ADAM or SGD.

In this section, we discuss the methods needed to understand how constraints can be represented, and define a novel diffusion process which projects the dynamics onto the submanifold defined by arbitrary sets of geometric constraints.
\subsection{Shake Algorithm}
The Shake algorithm takes as input a set of coordinates $x$ of a molecular system and a set of constraints $\sigma$. At each time step the coordinates
are updated according to the equations of motion (EOM) at hand (without constraint terms) and subsequently are corrected. In general, the EOM will lead to dynamics that do not
satisfy the constraints, and thus this correction is mandatory. 

Assuming masses of all the particles and delta time are unit we have the following equation for updating $x_i$ iteratively until the constraints are satisfied.
\begin{align}x_i^{(n)}= x_i^{(n-1)} - \sum_b{\lambda_b^{(n-1)}}\nabla\sigma_b(x_i)\end{align} where $x_i^{(n)}$ is the updated coordinate after n iterations of
satisfying constraints at each time step, $x_i$ is the initial coordinates at each time step, and $\lambda_b^{(n-1)}$ is the lagrange multiplier for each
constraint $\sigma_a$. The equation to solve at each iteration of each time step is
\begin{align} \sum_{\beta}{\lambda_{\beta}^{(n-1)}}A_{\alpha\beta}^{(n-1)}= \sigma_{\alpha}(x_i^{(n-1)})  \end{align}
with
\begin{align} A_{\alpha\beta}^{(n-1)}= \nabla\sigma_{\alpha}(x_i^{(n-1)}) \nabla\sigma_{\beta}(x_i).  \end{align}

The matrix $A^{(n-1)}_{\alpha\beta}$ is a symmetric matrix that describes how changes in particle positions affect both potential energy and constraint violations. The elements of the matrix are given by:

\begin{align}A^{(n-1)}_{\alpha\beta} = \frac{\partial^2 U}{\partial x_{\alpha} \partial x_{\beta}} + \sum_{k=1}^{N_c}\lambda^{(n-1)}_k\frac{\partial^2 \sigma_k}{\partial x_{\alpha} \partial x_{\beta}}\end{align}

where $N_c$ is the number of constraints. The matrix $A^{(n-1)}_{\alpha\beta}$ is used to solve for the Lagrange multipliers $\lambda^{(n)}_\beta$ , which are then used to adjust particle positions.
\subsection{Constraint-Induced Diffusion Process}

Suppose we want to incorporate a constraint, such as a distance constraint between two atoms. Let's denote this constraint by $f(x) = 0$ for simplicity. We can modify the diffusion process to satisfy this constraint by projecting the noise term onto the nullspace of the gradient of the constraint function, analagous to the $A$ matrix in Shake. This gives us:
$$dx = \sqrt{2D} (I - \nabla f(x) (\nabla f(x))^T) dB - D \nabla \log p_t(x) dt$$ where $D$ is the diffusion constant, $B$ is a standard Brownian motion, and $\nabla \log p_t(x)$ is the gradient of the log-probability density, which is equivalent to the negative of the potential energy function of the system.
Here, $I$ is the identity matrix, and $\nabla f(x) (\nabla f(x))^T$ is the outer product of the gradient of the constraint function, which represents the direction in which the constraint is changing. This projection ensures that the noise term does not push the system out of the constraint-satisfying space.

The covariance matrix of the perturbed Gaussian distribution of the denoising process can be understood formally using the Schur complement method, available in the Appendix. The key takeaway is the relation between constraints and correlations via projecting out the constraints in the Covariance matrix of a Multivariate Gaussian. This modified covariance matrix then defines the perturbed Gaussian distribution from which we can sample at each time step of the diffusion process. This is a good approximation when the constraints are nearly linear or when the changes in the variables are small. One note is that in if the projection operator is non-linear than the the process is no longer Gaussian, but since we deal with linearized constraints, or small changes at each time step, this is negligible as seen in the original Shake formalism. However, the Schur Complement method gives a more general formalism to ensure Gaussian-ness. 
\subsection{ Constraints as Correlations }
Consider, for instance, a scenario involving pairwise distance constraints between a set of variables denoted as $\boldsymbol{d} = { d_{ij} }$, where $d_{ij}$ signifies the distance separating variables $i$ and $j$. These constraints can be mathematically expressed through the set of functions $C_{ij}(\boldsymbol{\epsilon}) = ||\boldsymbol{\epsilon}_i - \boldsymbol{\epsilon}_j|| - d{ij} = 0$, which is applicable to all corresponding variable pairs $(i, j) \in \boldsymbol{d}$, influencing the samples drawn from a Multivariate Normal distribution.

The introduction of these geometric constraints essentially interrelates variables that were initially independent in the Gaussian distribution. In order to comprehend the implications of these constraints, the covariance matrix $\boldsymbol{\Sigma}'$ of the perturbed distribution $p'(\boldsymbol{\epsilon}')$ is worth examining:

\begin{equation}
\boldsymbol{\Sigma}' = \mathbb{E}_{\boldsymbol{\epsilon}' \sim p'} [\boldsymbol{\epsilon}' (\boldsymbol{\epsilon}')^T] - \mathbb{E}_{\boldsymbol{\epsilon}' \sim p'} [\boldsymbol{\epsilon}'] \mathbb{E}_{\boldsymbol{\epsilon}' \sim p'} [\boldsymbol{\epsilon}']^T,
\end{equation}

Here, the expectations are calculated over the perturbed distribution. The covariance matrix $\boldsymbol{\Sigma}'$ elucidates the correlations among variables that emerge as a result of the geometric constraints.

Importantly, these correlations, which are encoded within the covariance matrix of a multivariate Gaussian distribution, represent the constraints in the distribution. This provides a way to naturally incorporate constraint-based information into the model.

\subsection{Training and Sampling Algorithms }
\begin{figure*}[htb]

    \centering
    \begin{subfigure}{0.24\textwidth}
        \includegraphics[width=\linewidth]{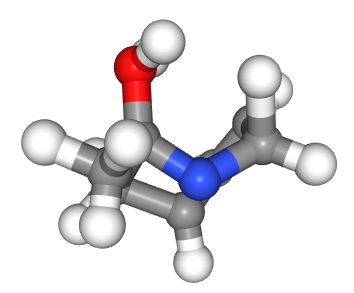}
        \label{fig:img0}
    \end{subfigure}%
    \hfill
    \begin{subfigure}{0.24\textwidth}
        \includegraphics[width=\linewidth]{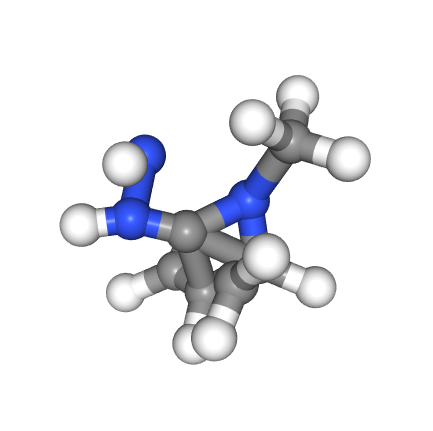}
        \label{fig:img2}
    \end{subfigure}%
    \hfill
    \begin{subfigure}{0.22\textwidth}
        \includegraphics[width=\linewidth]{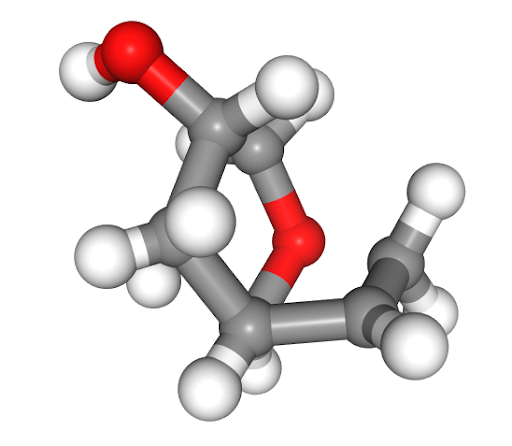}
        \label{fig:img4}
    \end{subfigure}%
    \hfill
    \begin{subfigure}{0.2\textwidth}
        \includegraphics[width=\linewidth]{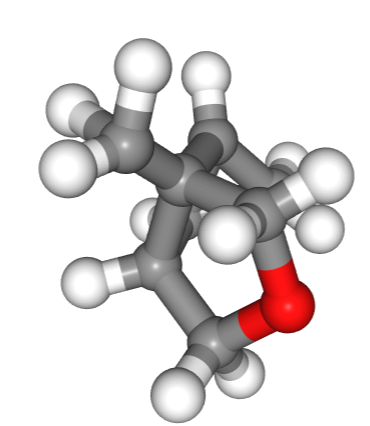}
        \label{fig:img6}
    \end{subfigure}
    
    \vskip\baselineskip

    \begin{subfigure}{0.24\textwidth}
        \includegraphics[width=\linewidth]{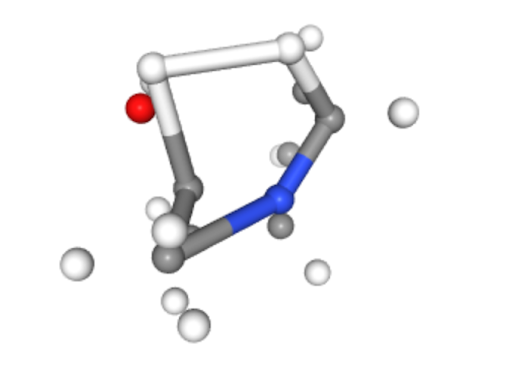}
        \label{fig:img1}
    \end{subfigure}%
    \hfill
    \begin{subfigure}{0.24\textwidth}
        \includegraphics[width=\linewidth]{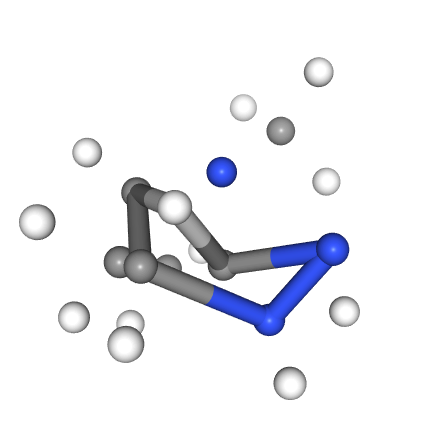}
        \label{fig:img3}
    \end{subfigure}%
    \hfill
    \begin{subfigure}{0.24\textwidth}
        \includegraphics[width=\linewidth]{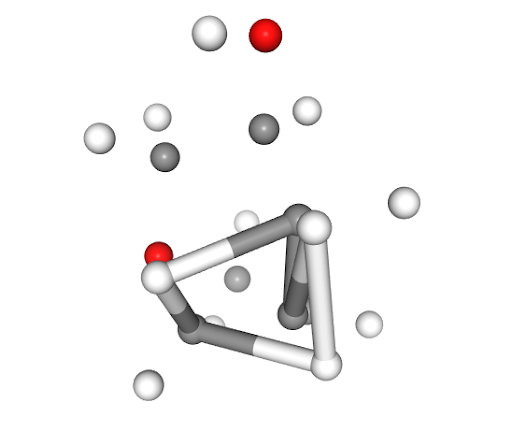}
        \label{fig:img5}
    \end{subfigure}%
    \hfill
    \begin{subfigure}{0.24\textwidth}
        \includegraphics[width=\linewidth]{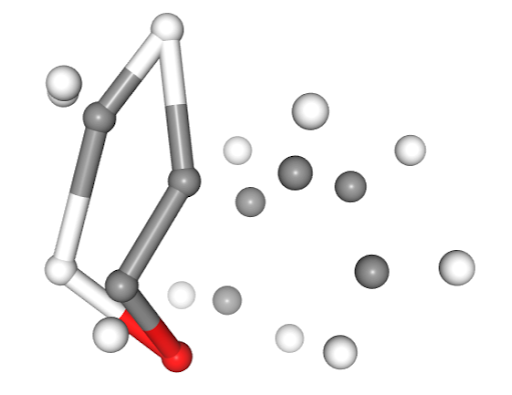}
        \label{fig:img7}
    \end{subfigure}
    \caption{Molecules generated with 6 atom cyclic constraints between 1.3-1.5 Angstroms each with bounds of .1 Angstrom. Atom types are generated as well, so we can not arbitrarily encode constraints between specific types of atoms in our current implementation, but this will be possible in further developments. }
\end{figure*}

\subsubsection{Training Process}
\begin{algorithm}[tb]
   \caption{Pseudo-Code for Training }
   \label{alg:example}
\begin{algorithmic}
    \STATE $t \sim U(0, T)$, $\epsilon \sim N(0, I)$
    \STATE Subtract center of gravity of coordinates from $\epsilon$: $\hat{\epsilon} = [\epsilon(x), 0] - [x, 0]$ 
    \STATE Compute $z_t = \alpha_t[x, h] + \sigma_t \hat{\epsilon}$ 
    \STATE Update $z_t \rightarrow x + \epsilon_s$, where $\epsilon_s = \text{Shake}(z_t ) - \alpha_tx$ 
    \STATE Compute $\epsilon_s' = \text{Shake}(\varphi(z_t ) + z_t ) - z_t  $ 
    \STATE Minimize $\mathcal{L}_c = |\epsilon_s - \epsilon_s'|_2^2$ 

\end{algorithmic}

\end{algorithm}

During training, in Algorithm 2, we first sample a time step $t$ and noise vector $\epsilon$ from uniform and Gaussian distributions respectively. Then subtract the center of gravity from the noise vector to ensure that it lies on a zero center of gravity subspace. Then compute the latent variable $z_t$ by scaling and adding the input coordinates $[x,h]$ with the noise vector. Finally, minimize the difference between the estimated noise vector and output of the neural network to optimize EDM. For each molecule between 5 and 15 constraints are sampled from $x$ for each batch element. The constraints are uniformly sampled from the pairs, triples, and quadruplets of the atom set of each molecule. This adds an extra layer of complexity due to the constraint distribution which we need to sample from the true data distribution.
\subsubsection{Generative Process}

In this generative process, we first sample a latent variable $z_T$ from a Gaussian distribution. Then iterate backwards through time and sample noise vectors $\epsilon$ at each step. Subtract the center of gravity of the coordinates from the noise vector to ensure that it lies on a zero center of gravity subspace. Then compute the latent variable $z_s$ by scaling and adding the input coordinates with the noise vector and previous latent variable. Finally, sample the input coordinates $[x,h]$ from a conditional distribution given the initial latent variable $z_0$. The Shake algorithm enforces the constraints, as in training, at each sampling step during generation. 

\section{Experiments }

In the experimental section of our study, we evaluate our proposed method by generating molecules with cyclic constraints in Figure 1. The cyclic constraints impose specific geometric relationships among atoms in a molecule, such as the bond distances, bond angles, and torsional angles, which are essential for maintaining the chemical stability and physical plausibility of the generated molecules. 

During the training phase, constraints are sampled from the dataset. This approach encourages the model to learn the distribution of constraints inherent in the training data, which reduces the Kullback-Leibler (KL) divergence between the data distribution and the model distribution. Consequently, the KL divergence during training is always minimized, promoting the model to generate molecules that closely resemble those in the training set.

For the practical implementation of this training procedure, we began with a pre-trained model provided by Welling et al.Our methodology then fine-tuned this pre-existing model using our constraint projection method. Due to time considerations and simplicity, our training and experiments focused on molecules consisting of 21 atoms. 

\section{Discussion}
 Our method serves as a potent tool for incorporating complex constraints in denoising diffusion processes, specifically when dealing with multi-constraint specifications. Its iterative nature allows it to address nonlinear constraint problems and extends the power of denoising diffusion probabilistic models to work with constraints. Thus allowing these models to leverage the structure inherent in many physical systems. Indeed, many of these systems come with prior structural knowledge, including geometric information like distances, torsions, bond angles, and generalizeable to other piece-wise polynomial terms. Such information can significantly enhance the training process and enable explicit sampling of subsets of the state space.

Although constraints can guide generation towards more physically plausible structures, there can be potential instability in the generation process. This instability may originate from discrepancies between constraints used during training and those applied during generation. It underlines the need for further work to establish robust training procedures that align more closely with the generation constraints. Especially, with application focused studies like generating peptides or ligands with specific interaction profiles. 

Though the language of our work is steeped in the semantics of Molecular Generation, the way we use geometric constraints to guide sampling mirrors a more general need of generative models in ML, which must navigate complex, structured probability spaces. 

Further exploration could include adapting our methodology to discern constraints intrinsically or applying it to optimization processes like gradient-based learning and potentially lead to more efficient or robust learning algorithms.

\printbibliography

@article{hoffmann2019generating,
  title={Generating Valid Euclidean Distance Matrices},
  author={Hoffmann, M and Noé, F},
  journal={arXiv preprint arXiv:1910.03131},
  year={2019},
  url={https://arxiv.org/abs/1910.03131}
}

@article{noe2019boltzmann,
  title={Boltzmann Generators: Sampling Equilibrium States of Many-Body Systems with Deep Learning},
  author={Noé, Frank and Olsson, Simon and Köhler, Jonas and Wu, Hao},
  journal={Science},
  volume={365},
  issue={6457},
  year={2019},
  publisher={American Association for the Advancement of Science},
  doi={10.1126/science.aaw1147},
  issn={0036-8075},
  eissn={1095-9203}
}

@inproceedings{satorras2021equivariant,
  title={E (n) Equivariant Graph Neural Networks},
  author={Satorras, Victor Garcia and Hoogeboom, Emiel and Welling, Max},
  booktitle={International Conference on Machine Learning},
  pages={9323--9332},
  year={2021}
}

@inproceedings{hoogeboom2023equivariant,
  title={Equivariant Diffusion for Molecule Generation in 3D},
  author={Hoogeboom, Emiel and Satorras, Victor Garcia and Vignac, Clement and Welling, Max},
  booktitle={International Conference on Machine Learning},
  pages={8867--8887},
  year={2023},
  url={https://arxiv.org/pdf/2203.17003.pdf}
}

@article{elber2011shake,
  title={SHAKE Parallelization},
  author={Elber, R. and Ruymgaart, A. P. and Hess, B.},
  journal={The European Physical Journal Special Topics},
  volume={200},
  pages={211--223},
  year={2011}
}

@article{chen2018neural,
  title={Neural Ordinary Differential Equations},
  author={Chen, Ricky T. Q. and Rubanova, Yulia and Bettencourt, Jesse and Duvenaud, David},
  journal={arXiv preprint arXiv:1806.07366},
  year={2018},
  url={https://arxiv.org/abs/1806.07366}
}

@article{ho2020denoising,
  title={Denoising Diffusion Probabilistic Models},
  author={Ho, Jonathan and Jain, Ajay and Abbeel, Pieter},
  journal={arXiv preprint arXiv:2006.11239},
  year={2020},
  url={https://arxiv.org/abs/2006.11239}
}

@inproceedings{corso2023diffdock,
  title={DiffDock: Diffusion Steps, Twists, and Turns for Molecular Docking},
  author={Corso, Gabriele and Stärk, Hannes and Jing, Bowen and Barzilay, Regina and Jaakkola, Tommi},
  booktitle={International Conference on Learning Representations},
  year={2023},
  url={https://arxiv.org/abs/2210.01776}
}

@inproceedings{jing2023eigenfold,
  title={EigenFold: Generative Protein Structure Prediction with Diffusion Models},
  author={Jing, Bowen and Erives, Ezra and Pao-Huang, Peter and Corso, Gabriele and Berger, Bonnie and Jaakkola, Tommi},
  booktitle={International Conference on Learning Representations, Machine Learning and Data Driven Discovery workshop},
  year={2023},
  url={https://arxiv.org/abs/2304.02198}
}

\section{Appendix A: Generalized Schur Complement for Multiple Constraints}

To obtain a generalized approach of Schur Complement for multiple distance constraints, let's consider a set of $M$ pairwise constraints between atoms. We can express each constraint as a function of the positions of the corresponding atoms:

\begin{equation}
f_m(\mathbf{x}_i, \mathbf{x}_j) = ||\mathbf{x}_i - \mathbf{x}_j||^2 - d_{ij}^2 = 0, \quad m = 1, 2, \ldots, M,
\end{equation}

where $d_{ij}$ is the distance constraint between atoms $i$ and $j$.

To incorporate all the constraints, we can form the combined gradient and Hessian matrices by stacking the corresponding matrices for each constraint:

\begin{equation}
\nabla \mathbf{f} = \begin{bmatrix}
\nabla f_1 \
\nabla f_2 \
\vdots \
\nabla f_M
\end{bmatrix},
\end{equation}

\begin{equation}
\nabla^2 \mathbf{f} = \begin{bmatrix}
\nabla^2 f_1 \
\nabla^2 f_2 \
\vdots \
\nabla^2 f_M
\end{bmatrix}.
\end{equation}

To project the Gaussian distribution with the original covariance matrix $\boldsymbol{\Sigma}$ onto the space of distance constraints, we can use the following generalized Schur complement:

\begin{equation}
\boldsymbol{\Sigma}' = \boldsymbol{\Sigma} - \boldsymbol{\Sigma} \nabla^2 \mathbf{f}^T (\nabla^2 \mathbf{f} \boldsymbol{\Sigma} \nabla^2 \mathbf{f}^T)^{-1} \nabla^2 \mathbf{f} \boldsymbol{\Sigma}.
\end{equation}
While the Schur complement method can be implemented iteratively for non-linear systems, it is computationally intensive due to the inversion of the Hessian matrix. However, it serves as an excellent theoretical tool, providing a precise representation of how constraints can be formally incorporated into the diffusion process.
On the other hand, the Schur complement method provides a direct way to project the covariance matrix of the atomic positions onto the space that satisfies the distance constraints. It essentially modifies the covariance matrix in a way that embeds the constraints, without needing to adjust the atomic positions. This approach formally modifies the probability distribution of interest, and may be more useful for theoretic insight.
\section{ Appendix C: Nonholonomic Constraints}

We are more interested in nonholonomic constraints where each constraint has possibly a lower and upper bound. As we mentioned earlier,
by adding a slack variable one can translate the nonholonomic constraints to holonomic ones. To formalize this, one sees that a constraint having
a lower and upper bound will either be completely satisfied or fail to satisfy a single boundary. Thus, we only have to consider
at most one holonomic constraint at each call to $Shake$ meaning each constraint with a lower and upper bound may be replaced by a lower, upper,
or no bound for each call.
\\ \\
To calculate the slack variable $y$ from $\sigma_{jk}:=\lVert x^{l}_i-x^{l}_j \rVert - d_{jk}$ which is $\leq or \geq 0$, one has
\begin{equation}
  y=\left\{
  \begin{array}{@{}ll@{}}
    max(0,||x^{l}_i-x^{l}_j||-d^u_{jk}), & \text{if}\ \leq \\
    max(0,d^l_{jk}-||x^{l}_i-x^{l}_j||), & \text{if}\ \geq
  \end{array}\right.
\end{equation}
where $d_{jk}$ is the lower or upper bound in case of nonholonomic constriants and the defined constraint
value for holonomic constraints. 

In the generative process, we define the initial values of $d_{jk}$ such that the constraints have little effects. The constraints are then linearly interpolated throughout the ODE until the predetermined boundary values of $d_{jk}$ are reached.

 \section{Appendix B: Incorporation of Logical Operators in Geometric Constraints}

The application of logical operators such as'AND', 'OR' and 'NOT' within geometric constraints enables a more flexible and representative modeling of physical and chemical systems. Real-world scenarios frequently require the satisfaction of multiple constraints following complex logical rules. Below, we detail the basic implementation of 'OR' and 'NOT' logical operators within the geometric constraints of our diffusion process while noting that the 'AND' operator is the basis of the formalism:

\subsection{'OR' Logic}

The 'OR' condition necessitates that at least one of two (or more) constraints be met. Let's denote two constraint functions as $f_1(x)$ and $f_2(x)$. The 'OR' logic can be integrated by constructing a composite constraint function that is satisfied when any of its constituent constraints is met. We can express this as:

\begin{equation}
g(x) = \min(f_1(x), f_2(x))
\end{equation}

In this case, if either $f_1(x) = 0$ or $f_2(x) = 0$ (or both), $g(x) = 0$, thereby meeting the 'OR' condition. Alternatively, we can employ a product of the constraints:

\begin{equation}
g(x) = f_1(x) \cdot f_2(x)
\end{equation}

If either $f_1(x) = 0$ or $f_2(x) = 0$ (or both), $g(x) = 0$, again adhering to the 'OR' logic. This method requires that both $f_1(x)$ and $f_2(x)$ are always non-negative.

\subsection{'NOT' Logic}
The "NOT" operator in the context of geometric constraints could be defined using the following equations. Let's say we have a constraint $f(x) = 0$. We want to define a NOT operator for this constraint. We can then define "NOT f(x)" as regions where f(x) does not equal zero, which can be represented with two inequality constraints which can be combined via the 'OR' operator to designate the 'NOT' operator.

We denote $\epsilon$ as a small positive number, then "NOT f(x)" can be represented as:

\begin{equation}
g_1(x) = f(x) + \epsilon < 0
\end{equation}

\begin{equation}
g_2(x) = f(x) - \epsilon > 0
\end{equation}

In the equations above, we have defined two regions (when $f(x)$ is smaller than $-\epsilon$ and larger than $\epsilon$) where "NOT f(x)" is true, thus defining a NOT operator for our constraints. Note that these regions depend on the choice of $\epsilon$.
\end{document}